\documentclass[11pt]{article}

\PassOptionsToPackage{table}{xcolor}
\usepackage[final]{acl}

\usepackage{times}
\usepackage{latexsym}

\usepackage[T1]{fontenc}

\usepackage[utf8]{inputenc}

\usepackage{microtype}

\usepackage{inconsolata}

\usepackage[most]{tcolorbox}
\usepackage{amsmath}
\usepackage{amssymb}
\usepackage{booktabs}
\usepackage{multirow}
\usepackage{array}
\usepackage{graphicx}

\usepackage[table]{xcolor}
\definecolor{best}{HTML}{9ac4de}
\definecolor{second}{HTML}{F4C688}
\tcbuselibrary{listings,breakable,skins}

\definecolor{egipBoxBack}{HTML}{FAFAFA}
\definecolor{egipBoxFrame}{HTML}{C9CED6}
\definecolor{egipTitleBack}{HTML}{EEF1F4}
\definecolor{egipTitleText}{HTML}{26313D}
\definecolor{egipCodeText}{HTML}{111827}

\newtcblisting{promptlisting}[2][]{
  enhanced,
  breakable,
  listing only,
  listing options={
    basicstyle=\ttfamily\footnotesize\color{egipCodeText},
    breaklines=true,
    columns=fullflexible,
    keepspaces=true,
    showstringspaces=false,
    aboveskip=0pt,
    belowskip=0pt
  },
  colback=egipBoxBack,
  colframe=egipBoxFrame,
  colbacktitle=egipTitleBack,
  coltitle=egipTitleText,
  boxrule=0.35pt,
  arc=1.5pt,
  outer arc=1.5pt,
  left=7pt,
  right=7pt,
  top=7pt,
  bottom=7pt,
  title={#2},
  fonttitle=\bfseries\small,
  boxed title style={
    boxrule=0pt,
    sharp corners,
    colback=egipTitleBack,
    left=7pt,
    right=7pt,
    top=3pt,
    bottom=3pt
  },
  attach boxed title to top left={
    xshift=0pt,
    yshift=-0.5pt
  },
  before skip=8pt,
  after skip=10pt,
  segmentation style={draw=egipBoxFrame, line width=0.25pt},
  #1
}

\newtcblisting{jsonlisting}[2][]{
  enhanced,
  breakable,
  listing only,
  listing options={
    basicstyle=\ttfamily\scriptsize\color{egipCodeText},
    breaklines=true,
    columns=fullflexible,
    keepspaces=true,
    showstringspaces=false,
    aboveskip=0pt,
    belowskip=0pt
  },
  colback=egipBoxBack,
  colframe=egipBoxFrame,
  colbacktitle=egipTitleBack,
  coltitle=egipTitleText,
  boxrule=0.35pt,
  arc=1.5pt,
  outer arc=1.5pt,
  left=7pt,
  right=7pt,
  top=7pt,
  bottom=7pt,
  title={#2},
  fonttitle=\bfseries\small,
  boxed title style={
    boxrule=0pt,
    sharp corners,
    colback=egipTitleBack,
    left=7pt,
    right=7pt,
    top=3pt,
    bottom=3pt
  },
  attach boxed title to top left={
    xshift=0pt,
    yshift=-0.5pt
  },
  before skip=8pt,
  after skip=10pt,
  segmentation style={draw=egipBoxFrame, line width=0.25pt},
}

\title{EGIP: An Evidence-Grounded Interaction Protocol for Clinical Agents}

\author{
  \textbf{Zecheng Ren\textsuperscript{1}},
  \textbf{Ruichen Cong\textsuperscript{1}},
  \textbf{Yi Min Chen\textsuperscript{1}},
  \textbf{Jing Zhang\textsuperscript{1}} \\
  \textbf{Feiyang Chen\textsuperscript{2}},
  \textbf{Wutong Zhang\textsuperscript{2}},
  \textbf{Kaixiang Wang\textsuperscript{2}},
  \textbf{Zihan Wang\textsuperscript{2}} \\
  \textbf{Jiong Lou\textsuperscript{2}},
  \textbf{Jie Li\textsuperscript{2}},
  \textbf{Qun Jin\textsuperscript{1}}\thanks{Corresponding author.} \\[0.5ex]
  \textsuperscript{1}Waseda University, Tokorozawa, Japan \\
  \textsuperscript{2}Shanghai Jiao Tong University, Shanghai, China \\
  \texttt{\href{mailto:z.ren@ruri.waseda.jp}{z.ren@ruri.waseda.jp}}
}

\begin{document}
\maketitle

\begin{abstract}
Interactive clinical agents acquire evidence through patient dialogue and diagnostic examinations, yet their performance also depends on how the interaction environment exposes information, available actions, and examination results. When these conditions are implicit or inconsistently controlled, final-answer accuracy can conflate diagnostic reasoning with the effects of the interaction protocol. We propose the \textbf{E}vidence-\textbf{G}rounded \textbf{I}nteraction \textbf{P}rotocol (EGIP), which enforces role-specific information boundaries, exposes diagnostic actions without revealing their results, retrieves examination results from stored clinical records, and organizes acquired evidence for final decision-making. By separating evidence acquisition from terminal prediction, EGIP enables different decision strategies to operate on the same observations. We evaluate EGIP across two clinical-agent frameworks, five datasets, and six language models. EGIP improves diagnostic accuracy over native baselines across all primary comparisons. A clinician review further shows that the proportion of automatically scored correct responses jointly judged evidence-supported rises from 10.20\% to 36.55\%. These findings support the effectiveness of EGIP as a unified interaction protocol for improving clinical-agent performance through controlled evidence access and acquisition. Code is released at \url{https://github.com/Nislab-r/EGIP}.

\end{abstract}

\begin{figure*}[t]
    \centering
    \includegraphics[width=0.95\textwidth]{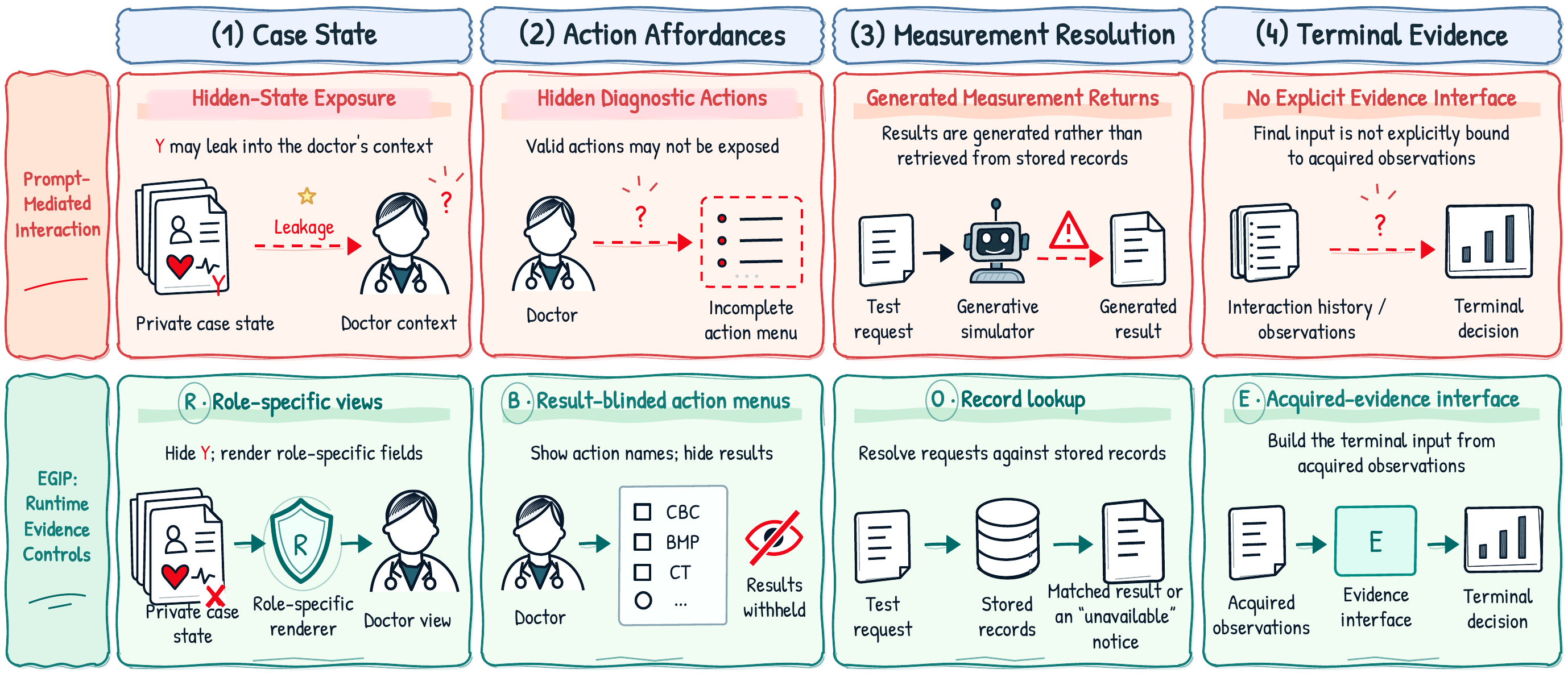}
    \caption{Evidence-access interfaces in clinical simulation. Top: potential issues in prompt-mediated interaction. Bottom: EGIP's role-specific views, result-blinded action menus, record lookup, and acquired-evidence interface. $Y$ denotes the reference target.}
    \label{fig:baseline_flaws}
\end{figure*}

\section{Introduction}

\begin{figure*}[t]
    \centering
    \includegraphics[width=0.98\textwidth]{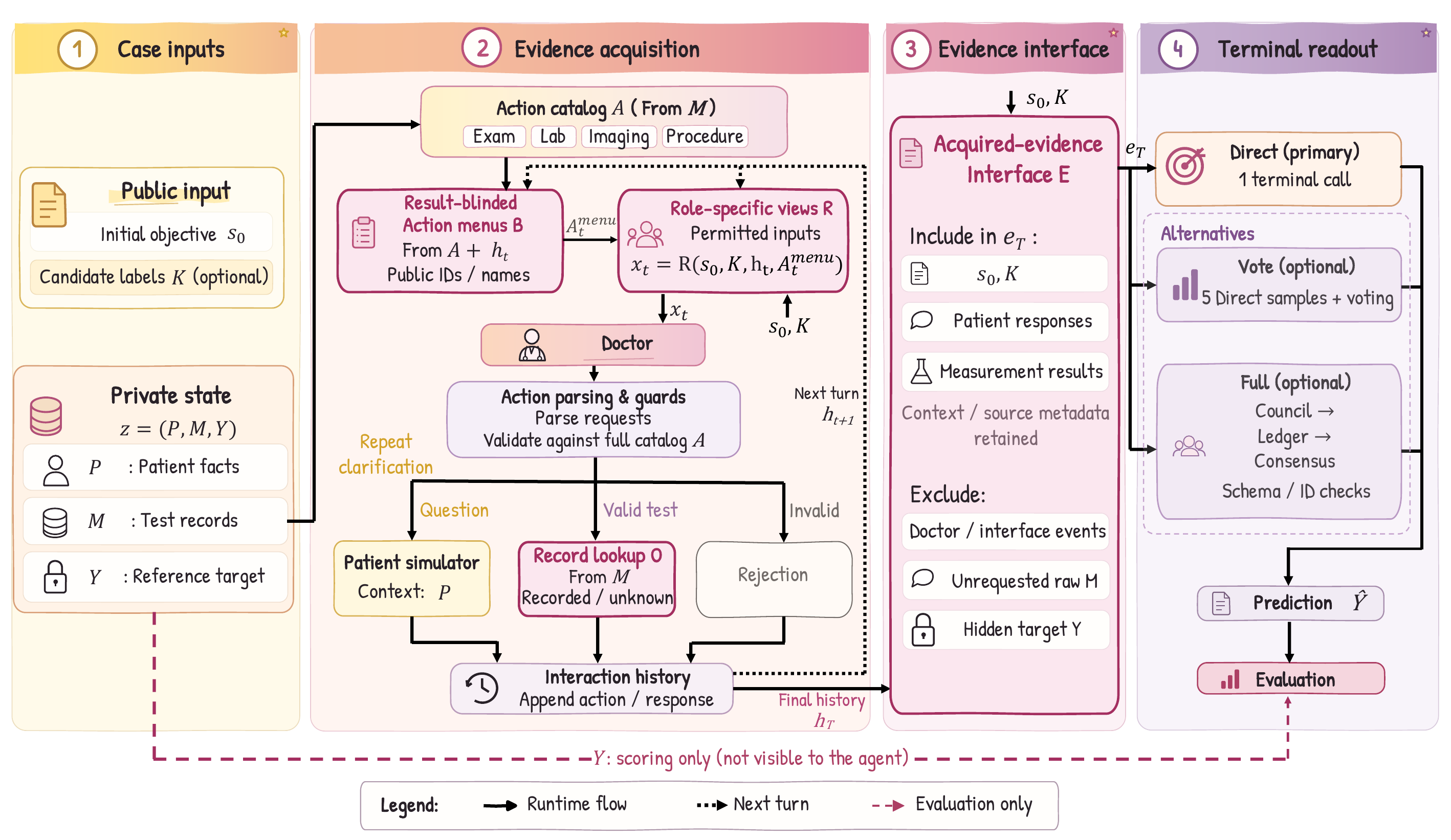}
    \caption{Overview of EGIP. Dark, bold labels and thick borders highlight the four core components: role-specific views ($R$), result-blinded action menus ($B$), record lookup ($O$), and the acquired-evidence interface ($E$). The protocol separates evidence acquisition from terminal prediction. The hidden reference target $Y$ is used only for post-prediction evaluation.}
    \label{fig:egip_framework}
\end{figure*}

Large language models (LLMs) have enabled clinical agents to conduct patient interviews, request diagnostic examinations, and make diagnoses through multi-turn interactions \citep{tu2025amie,schmidgall2026agentclinic,almansoori2025medagentsim}. Recent studies explore sequential symptom inquiry \citep{jia2025ddo,ren2025diallm}, medical tool use \citep{li2024mmedagent,shi2024ehragent}, and multi-agent clinical decision-making \citep{kim2024mdagents}. Unlike traditional medical question answering, interactive clinical agents must actively seek and acquire evidence before reaching a diagnosis. Their performance therefore depends not only on the underlying model's reasoning ability, but also on how the interaction environment determines what information is accessible, which diagnostic actions can be performed, and what observations are returned. The interaction environment thus plays a fundamental role in shaping the evidence available for clinical decision-making.

However, existing clinical-agent frameworks often rely on prompt-based role instructions and framework-specific simulation logic to manage evidence access \citep{schmidgall2026agentclinic,almansoori2025medagentsim}. Such mechanisms may expose information that has not been acquired through interaction, leave available diagnostic actions unclear, or return examination findings generated by a simulator rather than grounded in clinical records. For example, EvoPatient's pre-evolution patient agent leaks diagnoses and test results under broad history questions \citep{du2025evopatient}. AgentClinic's released doctor prompt gives a request format but no case-specific test catalog \citep{schmidgall2026agentclinic}, and AI Hospital's examiner is instructed to return ``No abnormalities'' for unrecorded tests \citep{fan2025aihospital}. Moreover, intermediate hypotheses and actually observed clinical evidence are not always clearly distinguished in the final decision process. These issues can alter the evidence available to an agent and consequently affect its diagnostic performance. A correct diagnosis may not reflect appropriate evidence acquisition, while an incorrect one may result from limitations of the interaction environment rather than diagnostic reasoning itself. This makes explicit control over clinical evidence access and acquisition essential for building reliable interactive clinical agents.

These problems suggest that evidence-grounded clinical interaction requires explicit controls over how information becomes available to an agent. We propose the \textbf{E}vidence-\textbf{G}rounded \textbf{I}nteraction \textbf{P}rotocol (EGIP) to control this evidence flow (Figure~\ref{fig:baseline_flaws}). EGIP first uses role-specific views to separate private case state from doctor-facing context, excluding undisclosed patient information and reference targets from the doctor's inputs. Its result-blinded action menus then make available diagnostic actions explicit without revealing their outcomes, allowing agents to plan evidence acquisition rather than guess the simulator's action vocabulary. When an examination is requested, record lookup retrieves its result from stored clinical records instead of generating plausible findings, preserving unavailable or unknown results. Finally, the acquired-evidence interface organizes observations returned during interaction for terminal prediction, treating candidate labels and agent assertions as hypotheses rather than independent clinical evidence. Together, these mechanisms establish a controlled and traceable path from information access and clinical actions to acquired evidence and final diagnosis.

We implement EGIP in two existing clinical-agent frameworks and evaluate it across six language models and five clinical datasets. EGIP-Direct, which uses a single-call terminal predictor, achieves higher mean accuracy across all primary comparisons. For example, with Qwen3-14B on MedAgentSim, accuracy increases from 34.58\% to 62.93\% on MedQA and from 63.08\% to 84.95\% on MIMIC-IV. Ablation studies further demonstrate the importance of result-blinded action menus and acquired clinical observations. An exploratory clinician review shows that the proportion of automatically scored correct outputs jointly judged evidence-supported increases from 10.20\% to 36.55\%. These findings demonstrate that controlling evidence access and acquisition can substantially improve clinical-agent performance without modifying the underlying language models.

Our main contributions are summarized as follows:
\begin{enumerate}
    \item We propose EGIP, a unified interaction protocol that regulates clinical evidence acquisition through role-specific views, result-blinded action menus, and record lookup. These mechanisms establish explicit controls over what information agents can access and how clinical evidence is obtained.
    \item We design a structured evidence interface that separates acquired clinical observations from agent-generated hypotheses and intermediate reasoning. This interface decouples evidence acquisition from final decision-making, allowing different diagnostic strategies to operate on the same evidence.
    \item We implement EGIP in two clinical-agent frameworks and evaluate it across six language models and five datasets. Extensive experiments demonstrate substantial improvements in diagnostic accuracy, while ablation studies and clinician assessments further validate the effectiveness of the protocol.
\end{enumerate}

\section{Related Work}

\subsection{LLM Agents and Evidence-Grounded Interaction}

LLM-agent research has expanded from isolated generation to interactive systems. ReAct, Toolformer, and Reflexion explore reasoning-action loops, tool use, and feedback-based refinement \citep{yao2023react,schick2023toolformer,shinn2023reflexion}. AutoGen and MetaGPT support coordinated multi-agent workflows \citep{wu2023autogen,hong2023metagpt}. AgentBench, WebArena, and OSWorld evaluate agents in diverse interactive environments \citep{liu2024agentbench,zhou2024webarena,xie2024osworld}. AgentBoard examines intermediate progress, while tau-bench measures reliability across repeated interactions \citep{ma2024agentboard,yao2024taubench}. These studies motivate examining outcomes alongside interaction conditions. EGIP focuses on evidence access and acquisition in clinical agents, specifying which information agents receive, which actions they can request, and how acquired observations inform final diagnosis.


\subsection{Medical Diagnostic Agents}

Medical LLM evaluation has expanded from static question answering to interactive diagnosis. PubMedQA and MedQA assess medical question answering \citep{jin2019pubmedqa,jin2021medqa}, while AMIE studies diagnostic dialogue \citep{tu2025amie}. AgentClinic and MedAgentSim simulate clinical encounters where doctor agents question patients and request examinations \citep{schmidgall2026agentclinic,almansoori2025medagentsim}. DiaLLM incorporates EHR information into test recommendation and diagnosis \citep{ren2025diallm}. MMedAgent and EHRAgent investigate tool use and reasoning over clinical records \citep{li2024mmedagent,shi2024ehragent}. MDAgents adapts collaboration to task complexity \citep{kim2024mdagents}. DDO separates symptom inquiry from diagnosis to optimize their objectives \citep{jia2025ddo}. EGIP separates evidence acquisition from terminal prediction, enabling comparisons of decision strategies on the same observations.

\section{Method}\label{sec:method}

\subsection{Problem Setting and Design Requirements}\label{sec:problem-setting}
We consider an interactive clinical-agent setting in which a doctor agent gathers information through patient questions and examination requests before producing a final answer. We define the private state $z$ of each case as:
\begin{equation}
z=(P,M,Y),
\label{eq:private-state}
\end{equation}
where $P$ contains patient-facing facts, $M$ stores examination records, and $Y$ is the reference target. The doctor initially receives a task objective $s_0$, including any initially disclosed case information, and optional candidate labels $K$. The reference target is hidden from agent-facing inputs and used only for post-prediction scoring. Candidate labels are possible answers rather than observed clinical evidence.

The history $h_t$ records the first $t$ acquisition rounds, with $h_0$ empty. The doctor selects its next action from the inputs available after those rounds. Patient questions are answered by a simulator conditioned on $P$. Valid examination requests are resolved against $M$. After acquisition, the initial objective, optional candidates, and acquired observations form a terminal input $e_T$. A decision strategy uses this input to produce $\hat{Y}$.

EGIP follows two design requirements. First, evidence acquisition requires explicit controls over visible information, requestable examinations, and returned results. Second, the final decision interface must distinguish acquired observations from candidate answers and agent-generated assertions. This allows different strategies to reuse the same evidence without repeating acquisition. Figure~\ref{fig:egip_framework} illustrates the components implementing these requirements.

\subsection{Controlled Evidence Acquisition}\label{sec:controlled-acquisition}
At each acquisition round, EGIP controls both the information presented to the doctor agent and the observations returned in response to its actions. This requires three complementary mechanisms: separating accessible information from private case state, exposing requestable diagnostic actions without disclosing their results, and resolving examinations against stored records. EGIP coordinates these mechanisms through a shared action–observation history.

\paragraph{Role-specific views.}
The renderer $R$ builds the doctor's input from the task objective, optional candidate labels, interaction history, and the current action menu. This input $x_t$ is defined as:
\begin{equation}
x_t=R(s_0,K,h_t,A_t^{\mathrm{menu}}),
\label{eq:doctor-input}
\end{equation} 
where $A_t^{\mathrm{menu}}$ denotes the current result-blinded action menu. The doctor-facing input excludes the reference target $Y$, unrequested measurement values in $M$, and patient facts in $P$ that have not been disclosed through interaction. The patient simulator instead receives the patient-facing facts and permitted dialogue context needed to answer the current question. These role-specific views are constructed from explicit rules for permitted fields, actions, and output formats, rather than relying on role instructions alone.


\paragraph{Result-blinded action menus.}
Without an explicit action interface, an agent must infer which examinations are available and how to request them. EGIP derives an action catalog $A$ from the case's stored examination records and constructs a prioritized, result-blinded menu $A_t^{\mathrm{menu}}=B(A,h_t)$. The menu exposes public test identifiers and names, excludes already requested tests, and does not reveal their findings. The displayed menu is a subset of $A$. Requested tests are validated against the complete catalog, not only the displayed subset. This makes action discovery explicit while keeping examination outcomes hidden until requested.


\paragraph{Record lookup and action handling.}

The runtime parses and validates each proposed action, replaces repeated valid test requests and explicit premature diagnosis markers with fixed clarification questions, and routes the remaining actions by type. Patient questions receive simulator-generated responses conditioned on $P$ and the permitted dialogue context. Let $a_t$ denote the routed action or rejection outcome, and let $r_t$ denote the corresponding response. Valid examination requests are resolved by $O(M,a_t)$, which returns the stored result with its available identifier, source, and availability metadata. Absent or unknown values are not replaced by generated findings. Invalid requests produce interface feedback rather than clinical observations. Each action and response is appended to the history:
\begin{equation}
h_{t+1}=h_t\mathbin{\Vert}[(a_t,r_t)],
\label{eq:history-update}
\end{equation}
where $\Vert$ denotes ordered concatenation. The updated history becomes the basis for the next interaction and, after acquisition, the evidence interface in Section~3.3.



\subsection{Evidence Interface and Decoupled Diagnosis}\label{sec:evidence-diagnosis}
\paragraph{Acquired-evidence interface.}
Although EGIP controls evidence acquisition as described in Section~\ref{sec:controlled-acquisition}, the resulting interaction history contains more than clinical observations. Alongside patient responses and examination results, it records doctor actions, intermediate assertions, and interface feedback. Passing this history unchanged to the final predictor risks conflating observations returned by the environment with content produced during interaction. EGIP therefore constructs a dedicated terminal input after $T$ acquisition rounds as:
\begin{equation}
e_T = E(s_0,K,h_T).
\label{eq:evidence-interface}
\end{equation}
The interface $e_T$ contains three fields: the initial objective $s_0$, optional candidate labels $K$, and observations extracted from $h_T$. The observations consist of ordered Patient and Measurement events. They retain their question or request context and available source and availability metadata. Doctor and Interface events remain in $h_T$ for traceability but are excluded as independent observation entries. The reference target $Y$ and unrequested measurements from $M$ are not included. Candidate labels and doctor assertions retained within question context are hypotheses, not independent patient findings. Terminal instructions specify that unknown or unavailable results must not be treated as normal findings. The interface thus distinguishes acquired observations from other interaction content without assuming that every returned statement is clinically verified.


\paragraph{Decoupled terminal diagnosis.}
The acquired-evidence interface is independent of the final decision procedure. Let $F_\theta^\tau$ denote the complete terminal procedure for backbone $\theta$ and strategy $\tau$, including any selection or validation. Given $e_T$ from Eq.~\eqref{eq:evidence-interface}, the terminal prediction is
\begin{equation}
\hat{Y}=F_{\theta}^{\tau}(e_T),\quad
\tau\in\{\mathrm{Direct},\mathrm{Vote},\mathrm{Full}\}.
\label{eq:terminal-strategy}
\end{equation}
Direct, the primary strategy, makes one terminal model call to predict a diagnosis from $e_T$. Vote aggregates five Direct-style predictions. Full uses Council--Ledger--Consensus followed by output-schema and observation-reference checks. These strategies are alternatives rather than successive stages. Their additional generated opinions or ledgers do not constitute newly acquired observations. In Full, reference validation checks the admissibility of source IDs associated with the initial objective and acquired observations. It does not establish whether those observations clinically entail the diagnosis. By providing the same acquired information to interchangeable decision strategies, EGIP separates how evidence is obtained from how it is interpreted for a final decision.



\section{Experiments}\label{sec:experiments}

\subsection{Experimental Setup}\label{sec:experimental-setup}

\paragraph{Frameworks and datasets.}
We compare Native and EGIP-Direct in MedAgentSim \citep{almansoori2025medagentsim} and AgentClinic \citep{schmidgall2026agentclinic} on five text-only benchmarks: MedQA (107 cases), MedQA\_Ext (214), NEJM (15), NEJM\_Ext (120), and MIMIC-IV (288) \citep{jin2021medqa,johnson2023mimiciv}. Native retains each framework's original workflow.

\paragraph{Models.}
The primary backbones are Qwen2.5-VL-7B \citep{bai2025qwen25vl}, Llama3.1-8B \citep{grattafiori2024llama3}, and Qwen3-8B/14B \citep{yang2025qwen3}. EGIP-Direct uses the same evidence-acquisition procedure and payload construction across these backbones, followed by a single-call Direct predictor. Llama3.3-70B-AWQ \citep{grattafiori2024llama3} and Qwen2.5-72B-AWQ \citep{qwen2024qwen25} provide auxiliary checks on MedQA and MIMIC-IV.

\paragraph{Execution settings.}
Within each framework, methods share the backbone, decoding settings, and a 20-turn cap. EGIP uses a 16K context and $T=19$ acquisition rounds followed by terminal prediction. Each model call allows up to 200 output tokens. We run three stochastic repetitions per configuration. Temperatures are 0.2 for MedAgentSim and 0.05 for AgentClinic. Runs with 7B--14B backbones used an NVIDIA A6000 GPU and two Spark devices. The 70B/72B backbones ran on a single NVIDIA A100 GPU.

\paragraph{Metrics.}
Accuracy (Acc.) is the percentage of scheduled outputs automatically scored correct. Unscored outputs and non-answers remain in the denominator. We report mean $\pm$ sample standard deviation (SD) across repetitions. Answer coverage is the percentage of scheduled outputs containing an identifiable final answer. Paired 95\% confidence intervals (CIs) use a case-cluster bootstrap. Human-review metrics are described in Section~\ref{sec:human-review}.

\begin{table*}[t]
\centering
\scriptsize
\setlength{\tabcolsep}{3.5pt}
\resizebox{\textwidth}{!}{%
\begin{tabular}{llrrrrr}
\toprule
\textbf{Backbone} & \textbf{Method} & \textbf{MedQA} & \textbf{MedQA\_Ext} & \textbf{NEJM} & \textbf{NEJM\_Ext} & \textbf{MIMIC-IV} \\
\midrule
\multirow{4}{*}{Qwen2.5-VL-7B} & MedAgentSim & $29.91\pm7.99$ & $31.78\pm4.15$ & $24.44\pm13.88$ & $10.83\pm0.83$ & $57.41\pm0.40$ \\
 & \cellcolor{blue!6}+ EGIP-Direct & \cellcolor{blue!6}$\mathbf{57.94}\pm2.47$ & \cellcolor{blue!6}$\mathbf{55.61}\pm0.81$ & \cellcolor{blue!6}$\mathbf{77.78}\pm3.85$ & \cellcolor{blue!6}$\mathbf{45.00}\pm3.63$ & \cellcolor{blue!6}$\mathbf{87.62}\pm0.20$ \\
 & AgentClinic & $33.33\pm3.28$ & $34.58\pm2.47$ & $26.67\pm6.67$ & $12.78\pm3.94$ & $66.44\pm2.09$ \\
 & \cellcolor{blue!6}+ EGIP-Direct & \cellcolor{blue!6}$\mathbf{57.01}\pm1.87$ & \cellcolor{blue!6}$\mathbf{54.67}\pm0.00$ & \cellcolor{blue!6}$\mathbf{73.33}\pm0.00$ & \cellcolor{blue!6}$\mathbf{51.39}\pm2.10$ & \cellcolor{blue!6}$\mathbf{85.88}\pm1.22$ \\
\midrule
\multirow{4}{*}{Llama3.1-8B} & MedAgentSim & $31.15\pm3.28$ & $29.28\pm1.43$ & $13.33\pm6.67$ & $12.50\pm0.83$ & $50.35\pm1.04$ \\
 & \cellcolor{blue!6}+ EGIP-Direct & \cellcolor{blue!6}$\mathbf{59.81}\pm1.87$ & \cellcolor{blue!6}$\mathbf{65.26}\pm3.57$ & \cellcolor{blue!6}$\mathbf{53.33}\pm6.67$ & \cellcolor{blue!6}$\mathbf{47.50}\pm2.20$ & \cellcolor{blue!6}$\mathbf{78.12}\pm2.28$ \\
 & AgentClinic & $21.81\pm4.32$ & $25.23\pm1.24$ & $4.44\pm3.85$ & $8.61\pm0.96$ & $44.10\pm3.18$ \\
 & \cellcolor{blue!6}+ EGIP-Direct & \cellcolor{blue!6}$\mathbf{61.37}\pm2.16$ & \cellcolor{blue!6}$\mathbf{63.55}\pm1.40$ & \cellcolor{blue!6}$\mathbf{62.22}\pm13.88$ & \cellcolor{blue!6}$\mathbf{45.56}\pm1.27$ & \cellcolor{blue!6}$\mathbf{78.12}\pm0.69$ \\
\midrule
\multirow{4}{*}{Qwen3-8B} & MedAgentSim & $33.96\pm2.35$ & $31.46\pm1.43$ & $15.56\pm3.85$ & $7.22\pm1.27$ & $59.72\pm1.04$ \\
 & \cellcolor{blue!6}+ EGIP-Direct & \cellcolor{blue!6}$\mathbf{55.14}\pm0.93$ & \cellcolor{blue!6}$\mathbf{54.21}\pm1.40$ & \cellcolor{blue!6}$\mathbf{57.78}\pm3.85$ & \cellcolor{blue!6}$\mathbf{40.28}\pm0.48$ & \cellcolor{blue!6}$\mathbf{66.90}\pm1.40$ \\
 & AgentClinic & $42.06\pm4.95$ & $36.29\pm1.50$ & $11.11\pm13.88$ & $6.67\pm0.83$ & $64.70\pm1.97$ \\
 & \cellcolor{blue!6}+ EGIP-Direct & \cellcolor{blue!6}$\mathbf{54.83}\pm0.54$ & \cellcolor{blue!6}$\mathbf{53.27}\pm0.47$ & \cellcolor{blue!6}$\mathbf{53.33}\pm0.00$ & \cellcolor{blue!6}$\mathbf{40.56}\pm1.27$ & \cellcolor{blue!6}$\mathbf{67.01}\pm0.35$ \\
\midrule
\multirow{4}{*}{Qwen3-14B} & MedAgentSim & $34.58\pm1.87$ & $37.54\pm2.21$ & $13.33\pm6.67$ & $5.56\pm1.92$ & $63.08\pm3.50$ \\
 & \cellcolor{blue!6}+ EGIP-Direct & \cellcolor{blue!6}$\mathbf{62.93}\pm1.43$ & \cellcolor{blue!6}$\mathbf{62.31}\pm1.77$ & \cellcolor{blue!6}$\mathbf{84.44}\pm7.70$ & \cellcolor{blue!6}$\mathbf{52.22}\pm2.68$ & \cellcolor{blue!6}$\mathbf{84.95}\pm0.72$ \\
 & AgentClinic & $46.11\pm1.43$ & $46.88\pm3.44$ & $15.56\pm10.18$ & $10.00\pm3.63$ & $74.42\pm3.75$ \\
 & \cellcolor{blue!6}+ EGIP-Direct & \cellcolor{blue!6}$\mathbf{61.06}\pm0.54$ & \cellcolor{blue!6}$\mathbf{61.37}\pm0.27$ & \cellcolor{blue!6}$\mathbf{77.78}\pm10.18$ & \cellcolor{blue!6}$\mathbf{53.33}\pm0.83$ & \cellcolor{blue!6}$\mathbf{85.19}\pm0.72$ \\
\bottomrule
\end{tabular}
}
\caption{Accuracy (\%) across four backbones, reported as mean $\pm$ sample standard deviation over three repetitions. Framework rows denote Native. Shaded entries denote EGIP-Direct. Bold marks the higher mean within each Native--EGIP pair.}
\label{tab:main-results}
\end{table*}

\begin{table}[t]
\centering\footnotesize
\setlength{\tabcolsep}{3pt}
\begin{tabular*}{\linewidth}{@{}l@{\extracolsep{\fill}}rr@{}}
\toprule
\textbf{Method} & \textbf{MedQA} & \textbf{MIMIC-IV} \\
\midrule
\multicolumn{3}{@{}l}{\textit{Llama3.3-70B-AWQ}} \\
MedAgentSim & $66.04\pm1.95$ & \cellcolor{blue!6}$\mathbf{79.51}\pm2.84$ \\
\quad + EGIP-Direct & \cellcolor{blue!6}$\mathbf{73.83}\pm2.47$ & $72.69\pm2.56$ \\
AgentClinic & $70.09\pm5.68$ & \cellcolor{blue!6}$\mathbf{80.21}\pm1.51$ \\
\quad + EGIP-Direct & \cellcolor{blue!6}$\mathbf{76.01}\pm1.08$ & \cellcolor{blue!6}$\mathbf{80.21}\pm1.84$ \\
\midrule
\multicolumn{3}{@{}l}{\textit{Qwen2.5-72B-AWQ}} \\
MedAgentSim & $50.16\pm2.35$ & $73.38\pm2.81$ \\
\quad + EGIP-Direct & \cellcolor{blue!6}$\mathbf{65.11}\pm0.54$ & \cellcolor{blue!6}$\mathbf{82.64}\pm1.25$ \\
AgentClinic & $51.71\pm2.35$ & $73.96\pm1.93$ \\
\quad + EGIP-Direct & \cellcolor{blue!6}$\mathbf{65.42}\pm0.00$ & \cellcolor{blue!6}$\mathbf{84.14}\pm1.40$ \\
\bottomrule
\end{tabular*}
\caption{Auxiliary Native--EGIP-Direct comparisons on MedQA and MIMIC-IV. Values are automated accuracy (\%), reported as mean $\pm$ sample standard deviation over three repetitions. Shading and bold mark the higher mean within each pair, including ties.}
\label{tab:strong-validation}
\end{table}

\begin{table*}[!htbp]
\centering\small
\setlength{\tabcolsep}{4pt}
\begin{tabular}{llrrl}
\toprule
\textbf{Dataset} & \textbf{Variant} & \textbf{Acc. (\%)} & \textbf{Drop (pp)} & \textbf{Paired 95\% CI} \\
\midrule
\multicolumn{5}{l}{\textbf{Panel A.} \textit{Acquisition components}} \\
\rowcolor{blue!6}
MedQA & EGIP-Direct & $57.94\pm2.47$ & -- & -- \\
 & w/o action menu & $24.30\pm0.00$ & \textbf{33.64} & [24.30, 42.68] \\
 & Native measurement access & $54.83\pm1.95$ & 3.12 & [$-1.56$, 8.10] \\
\rowcolor{blue!6}
NEJM\_Ext & EGIP-Direct & $45.00\pm3.63$ & -- & -- \\
 & w/o action menu & $40.56\pm0.96$ & 4.44 & [$-1.67$, 10.83] \\
 & Native measurement access & $41.39\pm3.94$ & 3.61 & [$-1.39$, 8.61] \\
\midrule
\multicolumn{5}{l}{\textbf{Panel B.} \textit{Terminal inputs with fixed acquisition history}} \\
\rowcolor{blue!6}
MedQA & EGIP-Direct (rerun) & $56.39\pm1.43$ & -- & -- \\
 & w/o observations & $20.56\pm1.62$ & \textbf{35.83} & [26.17, 45.17] \\
\rowcolor{blue!6}
NEJM\_Ext & EGIP-Direct (rerun) & $45.56\pm2.55$ & -- & -- \\
 & w/o observations & $32.50\pm1.44$ & \textbf{13.06} & [5.83, 20.28] \\
 & w/o candidate labels & $15.00\pm2.20$ & \textbf{30.56} & [22.22, 38.89] \\
\bottomrule
\end{tabular}
\caption{Ablations with Qwen2.5-VL-7B on MedAgentSim. Accuracy (\%) is mean $\pm$ sample standard deviation over three repetitions. Drop is the shaded control minus the variant within each panel and dataset, in percentage points (pp). Bold drops have paired, pointwise 95\% CIs excluding zero. Panel B uses rerun controls. ``w/o'' denotes removal. MedQA has no candidate labels.}
\label{tab:mechanism-ablation}
\end{table*}

\subsection{Results}

\paragraph{Primary comparisons.}
EGIP-Direct achieves higher mean accuracy in all primary comparisons across four backbones, both frameworks, and five datasets (Table~\ref{tab:main-results}). The magnitude varies by configuration: the largest mean gains occur on NEJM, while Qwen3-8B shows smaller gains on MIMIC-IV. Within Qwen3, the 14B model generally gains more than the 8B model.

\paragraph{Larger backbones.}
Table~\ref{tab:strong-validation} extends evaluation to 70B/72B backbones on two datasets. Qwen2.5-72B-AWQ achieves higher mean accuracy with EGIP-Direct in all four comparisons. Llama3.3-70B-AWQ gains on MedQA in both frameworks, but declines on MedAgentSim/MIMIC-IV and is unchanged on AgentClinic/MIMIC-IV. These auxiliary results show that gains do not extend uniformly across larger backbones and tasks.

\subsection{Ablation Study}
We hold the Direct procedure fixed to examine EGIP's acquisition components and terminal inputs (Table~\ref{tab:mechanism-ablation}). Panel A reports acquisition interventions: removing the action menu supplied by $B$ or replacing record lookup $O$ with native measurement access. Each variant acquires its own evidence. Panel B removes observations or candidate labels $K$ from terminal inputs $e_T$, keeping the acquisition history fixed.

\paragraph{Acquisition components.}
Removing the action menu $B$ produces a large MedQA drop, demonstrating the contribution of explicit test options with the terminal procedure held fixed. The action-menu effect on NEJM\_Ext and the record-lookup effects on both datasets remain uncertain.

\paragraph{Terminal inputs.}
Removing observations lowers accuracy on both datasets, with both paired intervals excluding zero. Removing candidate labels $K$ also substantially lowers accuracy on NEJM\_Ext. Thus, both acquired observations and candidate context contribute to Direct's performance and should be specified when comparing clinical agents.

\begin{figure}[!htbp]
    \centering
    \includegraphics[width=\linewidth]{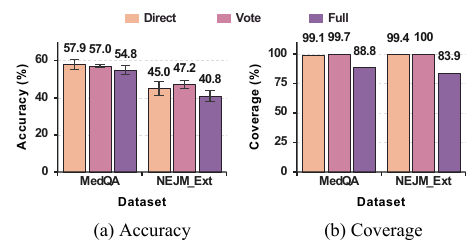}
    \caption{Comparison of Direct, Vote, and Full on the same acquired evidence for Qwen2.5-VL-7B on MedAgentSim. Accuracy and answer coverage are computed over all scheduled outputs.}
    \label{fig:terminal-comparison}
\end{figure}

\subsection{Terminal Strategies on Fixed Evidence}

The ablations hold the terminal procedure fixed. We next vary that procedure while holding acquired evidence fixed. Figure~\ref{fig:terminal-comparison} compares Direct, five-sample Vote, and Full for Qwen2.5-VL-7B on MedAgentSim using MedQA and NEJM\_Ext.

\paragraph{Direct as the Default.}
Direct achieves higher mean accuracy than Full on both datasets and retains nearly complete answer coverage. These results motivate our choice of Direct as the default terminal strategy in the evaluated settings. Full's lower coverage also highlights the need to evaluate answer production alongside accuracy.

\paragraph{Task-Dependent Voting.}
Relative to Direct, five-sample Vote modestly improves mean accuracy on NEJM\_Ext but not MedQA. Additional diagnosis samples therefore do not provide a uniform improvement, motivating task-specific evaluation of terminal strategies.

\begin{table*}[t]
\centering\small
\setlength{\tabcolsep}{4pt}
\begin{tabular}{lrrrl}
\toprule
\textbf{Support rule / dataset} & \textbf{Native, $n/N$ (\%)} & \textbf{Direct, $n/N$ (\%)} & \textbf{$\Delta$ (pp)} & \textbf{Paired 95\% CI} \\
\midrule
\multicolumn{5}{l}{\textbf{Panel A.} \textit{Correct-and-supported yield (539 complete pairs)}} \\
\rowcolor{blue!6}
\textbf{Both clinicians} $S$ \textbf{(CSY)} & 55/539 (\textbf{10.20}) & 197/539 (\textbf{36.55}) & \textbf{+26.35} & [20.93, 31.79] \\
Clinician 1 $S$ & 112/539 (20.78) & 259/539 (48.05) & +27.27 & [21.59, 33.09] \\
Clinician 2 $S$ & 74/539 (13.73) & 233/539 (43.23) & +29.50 & [23.89, 35.12] \\
At least one clinician $S$ & 131/539 (24.30) & 295/539 (54.73) & +30.43 & [24.72, 36.19] \\
\midrule
\multicolumn{5}{l}{\textbf{Panel B.} \textit{Evidence support within the shared correct-answer subset (105 pairs)}} \\
\rowcolor{blue!6}
\textbf{Both clinicians} $S$ & 43/105 (\textbf{40.95}) & 75/105 (\textbf{71.43}) & \textbf{+30.48} & [18.42, 43.16] \\
Clinician 1 $S$ & 85/105 (80.95) & 86/105 (81.90) & +0.95 & [-10.58, 12.84] \\
Clinician 2 $S$ & 58/105 (55.24) & 88/105 (83.81) & +28.57 & [18.69, 38.78] \\
At least one clinician $S$ & 100/105 (95.24) & 99/105 (94.29) & -0.95 & [-8.08, 5.36] \\
\midrule
\multicolumn{5}{l}{\textbf{Panel C.} \textit{Joint-support yield irrespective of correctness (both clinicians $S$)}} \\
\rowcolor{blue!6}
\textbf{All reviewed pairs} & 118/539 (\textbf{21.89}) & 231/539 (\textbf{42.86}) & \textbf{+20.96} & [15.21, 26.67] \\
MedQA & 92/422 (21.80) & 209/422 (49.53) & +27.73 & [20.95, 34.43] \\
NEJM\_Ext & 26/117 (22.22) & 22/117 (18.80) & -3.42 & [-14.53, 7.69] \\
\bottomrule
\end{tabular}
\caption{Exploratory review of Native and EGIP-Direct outputs. Panel A reports correct-and-supported yield. Panel B reports support in the post hoc shared correct-answer subset (105 pairs from 60 source cases). Panel C reports joint support irrespective of correctness. CSY requires automated correctness and two supported ($S$) ratings. $n/N$: qualifying outputs/complete pairs. $\Delta$: Direct minus Native in percentage points. CIs cluster by source case. Shading and bold mark joint-support summaries.}
\label{tab:human-support}
\end{table*}

\subsection{Human Review}
\label{sec:human-review}

The evidence interface records acquired observations but does not establish whether they support the final answer. We assess this relationship through clinician review.

\paragraph{Review setup.}
Two clinicians independently reviewed Native and EGIP-Direct outputs paired by case, backbone, framework, and repetition. We retain 539 complete pairs from 224 source cases after excluding nine incompletely rated pairs. The review covers Qwen2.5-VL-7B and Llama3.1-8B in both frameworks on MedQA, and Qwen2.5-VL-7B on MedAgentSim for NEJM\_Ext.

Reviewers judged each answer against its acquired evidence using a common rubric. Materials omitted reference answers and method labels, with a different randomized order for each reviewer. Repetition selection did not use outputs or scores. Ratings are supported ($S$), insufficiently supported ($U$), contradicted ($C$), indeterminate ($I$), or no final answer ($A$).

\paragraph{Evaluation metrics.}
For each method $m$ (Native or Direct), we define correct-and-jointly-supported yield (CSY) over all $N=539$ complete pairs as:
\begin{equation}
\mathrm{CSY}_{m}=\frac{100}{N}\sum_{j=1}^{N}c_j^{m}
\prod_{r=1}^{2}\mathbf{1}[\ell_{j,r}^{m}=S],
\label{eq:human-correct-support}
\end{equation}
where $j$ indexes complete pairs, $c_j^{m}$ equals 1 for outputs scored correct and 0 otherwise (including unscored outputs), $\ell_{j,r}^{m}$ is clinician $r$'s rating, and $\mathbf{1}[\cdot]$ is the indicator function.

Table~\ref{tab:human-support} reports three analyses. \textbf{Panel A} gives correct-and-supported yield under joint and alternative support rules. \textbf{Panel B} evaluates support among 105 pairs where both methods are scored correct and both clinicians identify an answer in each output. \textbf{Panel C} reports joint-support yield irrespective of correctness, overall and by dataset. Intervals use dataset-stratified source-case bootstrap resampling, retaining paired outputs and all model--framework records from each case.

\paragraph{Correct-and-Jointly-Supported Yield.}
Panel A shows higher CSY for Direct than Native (36.55\% vs.\ 10.20\%). The difference is $+26.35$ percentage points (95\% CI [20.93, 31.79]). Correct-and-supported yield is also higher under either clinician's ratings and the at-least-one-$S$ rule.

\paragraph{Support Among Shared Correct Answers.}
In the post hoc shared correct-answer subset (Panel B), Direct has higher joint support, with the interval excluding zero. The difference under Clinician 1's ratings is uncertain, and the at-least-one-$S$ rule shows no improvement.

\paragraph{Evidence support beyond accuracy.}
Joint support irrespective of correctness increases on MedQA but not clearly on NEJM\_Ext (Panel C). The review remains exploratory because agreement is low (37.44\%, Cohen's $\kappa=-0.007$ across 1,087 co-rated outputs, including outputs from incomplete pairs) and adjudication is incomplete.

\section{Discussion}
\paragraph{Evidence Access in Evaluation.}
The interface effects observed here motivate treating the interaction protocol as an experimental variable. Available actions and candidate context shape the task presented to an agent. Comparisons should therefore specify these conditions alongside the backbone and dataset. EGIP provides explicit points of control over evidence access and delivery. This allows evaluation to ask which interaction conditions contribute to performance, beyond reporting which system scores highest. Fixed-evidence comparisons further allow terminal strategies to be evaluated separately from evidence acquisition.

\paragraph{Evidence Support.}
Reporting correctness and support jointly distinguishes benchmark success from success with reviewer-endorsed support. Reporting support separately identifies supported answers that disagree with automated scores. These distinctions guide error analysis toward acquisition gaps, unsupported inference, or scoring disagreements. EGIP provides inspectable evidence for this analysis, while clinical review remains necessary to assess support.

\section{Conclusion}
We presented EGIP, a unified interaction protocol that controls evidence acquisition through role-specific views, result-blinded action menus, and record lookup, organizing acquired observations into a shared interface for diagnosis. This interface distinguishes observations from agent-generated hypotheses and enables different strategies to use the same evidence, decoupling acquisition from final decision-making. With this design, EGIP-Direct achieves higher mean accuracy in all primary comparisons, while clinician review finds a higher proportion of automatically scored correct outputs judged evidence-supported by both clinicians. These findings show that controlling evidence flow can improve clinical-agent performance without modifying the language models.

\section*{Limitations}
Larger-model evaluation covers only two datasets, limiting conclusions about generalization across model scales. Gains depend on the backbone and task, including a decline for Llama3.3-70B-AWQ on MedAgentSim/MIMIC-IV. Broader evaluation is needed to identify when EGIP benefits larger models.

Human review covers a subset of models and datasets and remains exploratory, with post hoc analyses, low agreement, and incomplete adjudication. The observed support gains therefore require further validation. Evidence interfaces cannot guarantee faithful patient simulation or clinically supported predictions, and these offline results do not establish clinical readiness.

\section*{Ethics Statement}
This study uses existing benchmarks and deidentified clinical resources for offline research. No new patient data were collected, and no reidentification was attempted. MIMIC-IV-derived data were used under applicable access and data-use requirements.

The clinician review was limited to evidence-support judgments on offline benchmark outputs, without patient contact or clinical intervention. These ratings are research annotations, not patient-specific medical advice or independent adjudication of diagnostic correctness. Joint support does not establish clinical safety or constitute clinician endorsement of deployment. EGIP is not intended for patient diagnosis or treatment. Clinical deployment would require independent validation, safety assessment, and human oversight.

AI assistance supported manuscript editing and checks against supplied analysis records. It did not generate experimental outcomes or adjudicate clinical correctness. The authors remain responsible for the manuscript, analyses, and disclosures.

\bibliography{references}

\clearpage
\appendix
\sloppy

\section{Experimental Details}\label{app:settings}

\subsection{Datasets, models, and generation}

\paragraph{Benchmarks and evaluation scope.}
We use the benchmark adaptations supplied by AgentClinic and MedAgentSim \citep{schmidgall2026agentclinic,almansoori2025medagentsim}, including cases derived from MedQA \citep{jin2021medqa} and MIMIC-IV \citep{johnson2023mimiciv}. The case counts in the experimental setup refer to these adapted evaluation sets. All inputs are text only, including those supplied to Qwen2.5-VL-7B.

\paragraph{Native and EGIP configurations.}
Native retains each framework's original protocol. EGIP-Direct combines EGIP acquisition with one terminal diagnosis call. Qwen3-8B and Qwen3-14B use BF16, non-thinking inference and a 16K context. The larger checkpoints are Llama3.3-70B-Instruct-AWQ and Qwen2.5-72B-Instruct-AWQ, abbreviated as Llama3.3-70B-AWQ and Qwen2.5-72B-AWQ. They use FP16 inference and a 16K context. Precision and model family differ across scales.

\paragraph{Generation and repetitions.}
The generation settings in the experimental setup apply to all three stochastic repetitions. The 200-token limit is per model call, not per case, and equal turn caps do not equate total model calls or tokens across systems. Direct--Native comparisons therefore evaluate complete workflows without matching inputs or total computation.

\paragraph{Action menus and observation metadata.}
Menu priorities use fixed rules over public test names without inspecting result values. Patient events retain the effective question in \texttt{question\_context}, without semantic filtering. Measurement events retain the canonical test request, source paths, recorded values, and availability labels. These correspond to \texttt{question\_context}, \texttt{source\_paths}, \texttt{recorded\_value}, and \texttt{result\_availability}. Availability is determined by matching the normalized whole value against predefined missing-value markers.

\subsection{Terminal-strategy implementation}\label{app:terminal-strategies}
Direct makes one terminal diagnosis call on $e_T$, requesting \texttt{DIAGNOSIS READY: <single diagnosis>}. Its prompt states that supplied acquired evidence is the factual basis, unknown findings are not normal, and candidate labels are hypotheses. Direct adds no generated ledger or source-ID checks. The one-call count excludes acquisition.

Vote uses five Direct samples on the same $e_T$ and selects the most frequent extracted diagnosis after formatting normalization, excluding abstentions and unparseable answers. Ties retain the first encountered label. Vote abstains if no eligible answer is available. The implementation reuses the corresponding Direct answer as its first sample.

Full applies Council--Ledger--Consensus to $e_T$. Council generates diagnostic proposals, Ledger organizes an evidence summary, and Consensus selects a diagnosis using these outputs and the original evidence. Schema and source-ID checks assess output structure, admissible observation references, and consistency with the proposed diagnosis. Unknown observations cannot serve as support or conflict references. At most one repair is allowed. Persistent validation failure or declared insufficiency produces abstention. These checks do not establish clinical entailment.

\subsection{Scoring and statistical procedures}\label{app:statistics}

\paragraph{Paired uncertainty.}
Paired 95\% intervals use 10,000 case-cluster bootstrap resamples, retaining paired methods and repetitions within each case. Intervals are pointwise and unadjusted, and quantify case-sampling uncertainty rather than variation across runs. Terminal-input ablations and human review use the procedures specified in their respective sections.

For the 28 Direct--Native comparisons listed in Table~\ref{tab:new-paired}, we additionally use 20,000 two-sided sign-flip randomizations of case-averaged paired differences with Holm adjustment across all 28 tests. These post hoc tests are exploratory. Because the CIs are pointwise and the tests are adjusted, a CI excluding zero need not correspond to an adjusted $p<0.05$. The shared-answer sensitivity procedure for Qwen2.5-VL-7B and Llama3.1-8B is detailed in Appendix~\ref{app:scoring}. Its conservative envelope combines sampling and unresolved-label uncertainty and is not an ordinary confidence interval.

\section{Scoring Sensitivity and Paired Comparisons}\label{app:scoring}
The reported accuracy counts only outputs scored correct. In a separate sensitivity analysis, we reuse consistent correctness labels for identical dataset--case--normalized-answer keys and treat absent or conflicting labels as unknown. Normalization removes formatting differences without merging clinical synonyms. All occurrences of an unresolved key share one unknown binary label. Minimizing and maximizing each method contrast over these labels yields finite-sample effect bounds. Repeating this calculation within case-cluster bootstrap resamples gives the conservative envelope. Because this procedure harmonizes labels across identical answers, its bounds need not contain the original automated-score difference. These bounds account for unresolved labels but not systematic scorer error.

Tables~\ref{tab:dn-ci} and \ref{tab:df-ci} report Direct--Native and Direct--Full contrasts for Qwen2.5-VL-7B and Llama3.1-8B. Ordinary intervals use automated scores, whereas effect bounds and conservative envelopes use the shared-answer sensitivity analysis. All intervals are pointwise and unadjusted.

\subsection{Score and output-boundary audits}\label{app:score-audits}
All Direct outputs in the 28 comparisons listed in Table~\ref{tab:new-paired} have automated scores and the requested \texttt{DIAGNOSIS READY:} marker. Some Native MedAgentSim outputs remain unscored and may still contain implicit diagnoses. Under the extreme assumption that all such outputs are correct, positive differences remain in 20 of 28 comparisons, although the aggregate MedAgentSim gains for Qwen3-8B and Qwen3-14B would reverse. These results show that automated scoring uncertainty affects the magnitude and interpretation of the reported gains.

\subsection{Paired comparisons for Qwen2.5-VL-7B and Llama3.1-8B}
Direct--Full comparisons use identical acquired evidence, whereas Direct--Native comparisons evaluate different acquisition workflows. Paired 95\% CIs are positive in all 20 Direct--Native comparisons, while some conservative envelopes include zero because of unresolved correctness labels. In both tables, Qwen denotes Qwen2.5-VL-7B and Llama3.1 denotes Llama3.1-8B. Differences are computed before rounding.

\begin{table*}[t]
\centering
\scriptsize
\setlength{\tabcolsep}{4pt}
\begin{tabular}{lllrrrr}
\toprule
\textbf{Model} & \textbf{Framework} & \textbf{Dataset} & \textbf{$\Delta$} & \textbf{Paired 95\% CI} & \textbf{Effect bounds} & \textbf{Conservative envelope} \\
\midrule
Qwen & MAS & MedQA & 28.04 & [20.56, 35.83] & [1.25, 27.10] & [-6.54, 34.89] \\
Qwen & MAS & MedQA\_Ext & 23.83 & [18.22, 29.60] & [0.62, 24.14] & [-5.14, 29.60] \\
Qwen & MAS & MIMIC-IV & 30.21 & [26.16, 34.26] & [9.26, 26.97] & [4.98, 30.90] \\
Qwen & MAS & NEJM & 53.33 & [20.00, 80.00] & [15.56, 55.56] & [-13.33, 82.22] \\
Qwen & MAS & NEJM\_Ext & 34.17 & [26.39, 42.22] & [14.72, 40.56] & [6.94, 48.33] \\
Qwen & AC & MedQA & 23.68 & [15.26, 32.09] & [15.26, 28.04] & [6.85, 36.14] \\
Qwen & AC & MedQA\_Ext & 20.09 & [14.80, 25.70] & [12.62, 23.83] & [7.32, 29.28] \\
Qwen & AC & MIMIC-IV & 19.44 & [14.81, 23.96] & [15.74, 25.23] & [11.00, 29.75] \\
Qwen & AC & NEJM & 46.67 & [15.56, 73.33] & [26.67, 51.11] & [-4.44, 75.56] \\
Qwen & AC & NEJM\_Ext & 38.61 & [30.28, 47.22] & [17.50, 43.33] & [9.44, 51.94] \\
Llama3.1 & MAS & MedQA & 28.66 & [18.69, 38.32] & [15.58, 35.83] & [5.61, 44.55] \\
Llama3.1 & MAS & MedQA\_Ext & 35.98 & [29.75, 42.06] & [19.78, 45.02] & [13.24, 50.78] \\
Llama3.1 & MAS & MIMIC-IV & 27.78 & [22.22, 33.33] & [16.44, 34.61] & [10.88, 39.58] \\
Llama3.1 & MAS & NEJM & 40.00 & [24.44, 55.56] & [6.67, 51.11] & [-15.56, 68.89] \\
Llama3.1 & MAS & NEJM\_Ext & 35.00 & [27.50, 42.78] & [11.67, 41.11] & [4.44, 48.89] \\
Llama3.1 & AC & MedQA & 39.56 & [30.84, 47.98] & [28.97, 42.99] & [20.25, 51.40] \\
Llama3.1 & AC & MedQA\_Ext & 38.32 & [32.09, 44.70] & [25.86, 44.86] & [19.16, 50.93] \\
Llama3.1 & AC & MIMIC-IV & 34.03 & [28.47, 39.47] & [26.27, 39.35] & [20.49, 44.33] \\
Llama3.1 & AC & NEJM & 57.78 & [40.00, 73.33] & [33.33, 71.11] & [11.11, 86.67] \\
Llama3.1 & AC & NEJM\_Ext & 36.94 & [28.89, 45.00] & [24.44, 48.06] & [16.39, 55.83] \\
\bottomrule
\end{tabular}
\caption{Direct minus Native in percentage points. MAS: MedAgentSim. AC: AgentClinic. CIs are paired, pointwise 95\% intervals. Effect bounds are not confidence intervals.}
\label{tab:dn-ci}
\end{table*}

\begin{table*}[t]
\centering
\scriptsize
\setlength{\tabcolsep}{4pt}
\begin{tabular}{lllrrrr}
\toprule
\textbf{Model} & \textbf{Framework} & \textbf{Dataset} & \textbf{$\Delta$} & \textbf{Paired 95\% CI} & \textbf{Effect bounds} & \textbf{Conservative envelope} \\
\midrule
Qwen & MAS & MedQA & 3.12 & [-2.81, 9.35] & [-3.12, 8.41] & [-9.97, 14.33] \\
Qwen & MAS & MedQA\_Ext & 3.89 & [-0.31, 7.79] & [-1.56, 8.10] & [-5.76, 12.15] \\
Qwen & MAS & MIMIC-IV & 11.69 & [8.56, 15.05] & [8.56, 15.62] & [4.98, 19.33] \\
Qwen & MAS & NEJM & 22.22 & [6.67, 40.00] & [6.67, 22.22] & [-6.67, 40.00] \\
Qwen & MAS & NEJM\_Ext & 4.17 & [-2.22, 10.83] & [-3.89, 12.78] & [-10.83, 20.00] \\
Qwen & AC & MedQA & 6.23 & [-0.31, 12.77] & [0.62, 9.35] & [-6.54, 16.51] \\
Qwen & AC & MedQA\_Ext & 2.80 & [-2.02, 7.79] & [-2.80, 6.70] & [-7.79, 11.37] \\
Qwen & AC & MIMIC-IV & 9.84 & [6.37, 13.43] & [6.60, 13.43] & [2.89, 17.01] \\
Qwen & AC & NEJM & 15.56 & [0.00, 35.56] & [11.11, 20.00] & [-6.67, 40.00] \\
Qwen & AC & NEJM\_Ext & 13.06 & [5.83, 20.83] & [1.67, 20.00] & [-6.11, 27.50] \\
Llama3.1 & MAS & MedQA & 23.36 & [16.20, 30.53] & [17.13, 30.22] & [9.66, 38.01] \\
Llama3.1 & MAS & MedQA\_Ext & 28.04 & [22.90, 33.02] & [21.50, 34.11] & [16.67, 39.25] \\
Llama3.1 & MAS & MIMIC-IV & 37.27 & [33.22, 41.32] & [29.98, 39.70] & [25.58, 43.63] \\
Llama3.1 & MAS & NEJM & 13.33 & [-4.44, 31.11] & [-2.22, 24.44] & [-17.78, 37.78] \\
Llama3.1 & MAS & NEJM\_Ext & 21.11 & [14.17, 28.06] & [11.94, 27.22] & [5.56, 33.89] \\
Llama3.1 & AC & MedQA & 33.02 & [24.61, 41.43] & [23.99, 33.33] & [15.58, 41.74] \\
Llama3.1 & AC & MedQA\_Ext & 34.42 & [28.66, 40.03] & [23.21, 37.38] & [17.29, 43.15] \\
Llama3.1 & AC & MIMIC-IV & 36.81 & [32.29, 41.44] & [29.75, 39.00] & [25.12, 43.40] \\
Llama3.1 & AC & NEJM & 37.78 & [20.00, 55.56] & [6.67, 35.56] & [-6.67, 51.11] \\
Llama3.1 & AC & NEJM\_Ext & 21.67 & [14.17, 29.17] & [6.39, 23.33] & [-0.00, 29.72] \\
\bottomrule
\end{tabular}
\caption{Direct minus Full in percentage points. MAS: MedAgentSim. AC: AgentClinic. CIs are paired, pointwise 95\% intervals. Effect bounds are not confidence intervals.}
\label{tab:df-ci}
\end{table*}

\section{Acquisition and Terminal-Input Ablation Protocols}\label{app:ablation-protocol}

For the acquisition ablations in Table~\ref{tab:mechanism-ablation}, each variant acquires its own evidence after removing the action menu or replacing record lookup with native measurement access. The one-call Direct procedure is held fixed. These interventions change the acquisition workflow and its observations, so they do not isolate terminal reasoning.

For terminal-input ablations, we hold acquisition histories fixed and remove either observation entries or candidate labels from the Direct input. The remaining payload, prompt, and generation settings are held fixed. Because the serving runtime differs from that used for the earlier collection, full-input controls are rerun alongside these ablations. Controls from the two panels are not pooled. Paired intervals use 20,000 source-case bootstrap resamples, retaining all three repetitions. These interventions measure sensitivity to terminal inputs rather than an additive decomposition of the overall Native-to-EGIP improvement.

\section{Auxiliary Validation with Larger Backbones}\label{app:auxiliary}
Table~\ref{tab:strong-validation} reports auxiliary checks with Llama3.3-70B-AWQ and Qwen2.5-72B-AWQ on MedQA and MIMIC-IV. These experiments examine whether the main findings extend to larger backbones.

Qwen2.5-72B-AWQ has higher mean accuracy with EGIP-Direct in all four comparisons. Llama3.3-70B-AWQ has higher means on MedQA, but lower accuracy on MedAgentSim/MIMIC-IV and unchanged accuracy on AgentClinic/MIMIC-IV. Neither Llama MedQA increase is significant after Holm adjustment (Table~\ref{tab:new-paired}). These exploratory results support applicability across model scales, with gains depending on the backbone and task.

\section{Human-Review Protocol and Sensitivity}\label{app:human-protocol}
For the review underlying Table~\ref{tab:human-support}, source cases were shuffled with a fixed seed and assigned one of the three repetitions in rotation, without consulting outputs or scores. The same repetition was used across the included methods, models, and frameworks for each case. The initial sample contained 548 pairs from MedQA and NEJM\_Ext. Excluding nine incompletely rated pairs leaves 539 complete pairs. The final analysis includes 422 MedQA pairs and 117 NEJM\_Ext pairs from 224 source cases. The review covers Qwen2.5-VL-7B and Llama3.1-8B in both frameworks on MedQA, and Qwen2.5-VL-7B on MedAgentSim for NEJM\_Ext.

Without seeing reference answers, two clinicians independently assigned each output one of five ratings: supported ($S$), insufficiently supported ($U$), contradicted ($C$), indeterminate ($I$), or no final answer ($A$). Review items used method-masked identifiers and reviewer-specific randomized orders. Original evidence and answer text were preserved, so textual cues could still reveal the method. Doctor assertions in question context and candidate labels were not treated as patient facts. Unknown findings were not treated as normal findings. Disagreement adjudication was not completed.

Correct-and-jointly-supported yield combines automated correctness with two clinician $S$ ratings using the common denominator of 539 complete pairs. Unscored outputs do not count as correct, but are not treated as confirmed errors. Two $S$ ratings indicate joint support rather than adjudicated clinical correctness. We use 20,000 dataset-stratified source-case bootstrap resamples, retaining paired outputs and all model--framework records from each case. We report pointwise 95\% intervals without multiple-comparison adjustment.

Agreement is calculated over all 1,087 outputs rated by both clinicians, including co-rated outputs whose paired counterpart has incomplete ratings. Paired outcome analyses instead use the 1,078 outputs in 539 complete pairs. The shared correct-answer subset is selected post hoc. These analyses assess evidence support within the reviewed sample and do not establish clinician-verified diagnostic accuracy.

\section{NEJM\_Ext Task Strata}\label{app:targets}
NEJM\_Ext contains 90 cases asking for a diagnosis, 13 asking about etiology, pathogens, or mechanisms, and 17 with other targets. Because the runtime requests a diagnosis, this creates a potential target mismatch. We retain all 120 cases in the main analysis and report target-stratified results as a robustness check. For Qwen2.5-VL-7B and Llama3.1-8B, the 90 explicitly diagnostic cases retain positive Direct--Native conservative envelopes across all four model--framework configurations, indicating that gains for these models are not driven solely by non-diagnostic targets. Table~\ref{tab:strata} reports strata for all four primary backbones.

\begin{table*}[t]
\centering\scriptsize
\setlength{\tabcolsep}{3pt}
\resizebox{\textwidth}{!}{%
\begin{tabular}{lrllrrllr}
\toprule
\textbf{Stratum} & \textbf{Cases} & \textbf{Model} & \textbf{Framework} & \textbf{Native} & \textbf{Direct} & \textbf{Paired 95\% CI} & \textbf{Conservative envelope} & \textbf{Native $U$} \\
\midrule
Diagnosis & 90 & Qwen2.5-VL-7B & MAS & 13.70 & 45.93 & [23.33, 41.48] & [2.96, 48.89] & --- \\
Diagnosis & 90 & Qwen2.5-VL-7B & AC & 14.81 & 54.81 & [30.00, 49.63] & [8.89, 53.33] & --- \\
Diagnosis & 90 & Llama3.1-8B & MAS & 12.22 & 49.26 & [27.78, 46.30] & [4.81, 51.48] & --- \\
Diagnosis & 90 & Llama3.1-8B & AC & 6.67 & 45.93 & [29.63, 48.52] & [19.26, 58.89] & --- \\
Etiology etc. & 13 & Qwen2.5-VL-7B & MAS & 2.56 & 28.21 & [7.69, 46.15] & [2.56, 56.41] & --- \\
Etiology etc. & 13 & Qwen2.5-VL-7B & AC & 12.82 & 23.08 & [-5.13, 28.21] & [-12.82, 51.28] & --- \\
Etiology etc. & 13 & Llama3.1-8B & MAS & 12.82 & 35.90 & [2.56, 46.15] & [-20.51, 61.54] & --- \\
Etiology etc. & 13 & Llama3.1-8B & AC & 7.69 & 41.03 & [12.82, 56.41] & [-2.56, 74.36] & --- \\
Other & 17 & Qwen2.5-VL-7B & MAS & 1.96 & 52.94 & [29.41, 72.55] & [9.80, 70.59] & --- \\
Other & 17 & Qwen2.5-VL-7B & AC & 1.96 & 54.90 & [27.45, 76.47] & [7.84, 76.47] & --- \\
Other & 17 & Llama3.1-8B & MAS & 13.73 & 47.06 & [13.73, 52.94] & [-5.88, 54.90] & --- \\
Other & 17 & Llama3.1-8B & AC & 19.61 & 47.06 & [3.92, 50.98] & [-13.73, 56.86] & --- \\
\midrule
Diagnosis & 90 & Qwen3-8B & MAS & 8.15 & 41.85 & --- & --- & 61 \\
Etiology etc. & 13 & Qwen3-8B & MAS & 7.69 & 30.77 & --- & --- & 11 \\
Other & 17 & Qwen3-8B & MAS & 1.96 & 39.22 & --- & --- & 14 \\
Diagnosis & 90 & Qwen3-8B & AC & 7.04 & 42.22 & --- & --- & 0 \\
Etiology etc. & 13 & Qwen3-8B & AC & 12.82 & 30.77 & --- & --- & 0 \\
Other & 17 & Qwen3-8B & AC & 0.00 & 39.22 & --- & --- & 0 \\
Diagnosis & 90 & Qwen3-14B & MAS & 7.04 & 55.93 & --- & --- & 160 \\
Etiology etc. & 13 & Qwen3-14B & MAS & 2.56 & 43.59 & --- & --- & 30 \\
Other & 17 & Qwen3-14B & MAS & 0.00 & 39.22 & --- & --- & 34 \\
Diagnosis & 90 & Qwen3-14B & AC & 11.85 & 57.41 & --- & --- & 0 \\
Etiology etc. & 13 & Qwen3-14B & AC & 10.26 & 41.03 & --- & --- & 0 \\
Other & 17 & Qwen3-14B & AC & 0.00 & 41.18 & --- & --- & 0 \\
\bottomrule
\end{tabular}
}
\caption{NEJM\_Ext target strata for the four primary backbones. Values are automated accuracy (\%) over three repetitions. Paired 95\% CIs and conservative envelopes report Direct minus Native in percentage points for Qwen2.5-VL-7B and Llama3.1-8B. $U$ counts Native unscored outputs across repetitions for Qwen3-8B and Qwen3-14B. Dashes denote unreported statistics. All 120 cases are retained, with strata defined by question semantics.}
\label{tab:strata}
\end{table*}

\section{Paired Comparisons for Qwen3 and Larger Backbones}\label{app:new-comparisons}
Table~\ref{tab:new-paired} lists 28 Direct--Native comparisons: 20 for Qwen3-8B and Qwen3-14B across all five datasets, and eight for Llama3.3-70B-AWQ and Qwen2.5-72B-AWQ on MedQA and MIMIC-IV. All four backbones are evaluated in both frameworks. $\Delta$ is Direct minus Native accuracy in percentage points, and $U$ counts native unscored outputs across three repetitions. The $p$ values come from exploratory sign-flip tests with Holm adjustment across all 28 comparisons.

\begin{table*}[t]
\centering\scriptsize
\setlength{\tabcolsep}{3.5pt}
\begin{tabular}{lllrlrr}
\toprule
\textbf{Model} & \textbf{Framework} & \textbf{Dataset} & \textbf{$\Delta$} & \textbf{Paired 95\% CI} & \textbf{Holm $p$} & \textbf{$U$} \\
\midrule
Qwen3-8B & MAS & MedQA & +21.18 & [13.71, 28.97] & 0.0014 & 91 \\
Qwen3-8B & MAS & MedQA\_Ext & +22.74 & [16.98, 28.35] & 0.0014 & 182 \\
Qwen3-8B & MAS & MIMIC-IV & +7.18 & [2.08, 12.27] & 0.0496 & 97 \\
Qwen3-8B & MAS & NEJM & +42.22 & [15.56, 68.89] & 0.0812 & 8 \\
Qwen3-8B & MAS & NEJM\_Ext & +33.06 & [24.44, 41.67] & 0.0014 & 86 \\
Qwen3-8B & AC & MedQA & +12.77 & [4.98, 20.56] & 0.0295 & 0 \\
Qwen3-8B & AC & MedQA\_Ext & +16.98 & [11.06, 22.90] & 0.0014 & 0 \\
Qwen3-8B & AC & MIMIC-IV & +2.31 & [-3.01, 7.41] & 0.8372 & 0 \\
Qwen3-8B & AC & NEJM & +42.22 & [20.00, 64.44] & 0.0496 & 0 \\
Qwen3-8B & AC & NEJM\_Ext & +33.89 & [25.00, 42.78] & 0.0014 & 0 \\
Qwen3-14B & MAS & MedQA & +28.35 & [19.31, 37.07] & 0.0014 & 114 \\
Qwen3-14B & MAS & MedQA\_Ext & +24.77 & [19.47, 30.22] & 0.0014 & 223 \\
Qwen3-14B & MAS & MIMIC-IV & +21.88 & [17.59, 26.39] & 0.0014 & 183 \\
Qwen3-14B & MAS & NEJM & +71.11 & [53.33, 86.67] & 0.0028 & 27 \\
Qwen3-14B & MAS & NEJM\_Ext & +46.67 & [38.33, 55.00] & 0.0014 & 224 \\
Qwen3-14B & AC & MedQA & +14.95 & [7.17, 22.74] & 0.0060 & 0 \\
Qwen3-14B & AC & MedQA\_Ext & +14.49 & [9.03, 20.09] & 0.0014 & 0 \\
Qwen3-14B & AC & MIMIC-IV & +10.76 & [6.71, 14.93] & 0.0014 & 0 \\
Qwen3-14B & AC & NEJM & +62.22 & [44.44, 80.00] & 0.0045 & 0 \\
Qwen3-14B & AC & NEJM\_Ext & +43.33 & [35.28, 51.39] & 0.0014 & 0 \\
Llama3.3-70B-AWQ & MAS & MedQA & +7.79 & [1.25, 14.64] & 0.1212 & 0 \\
Llama3.3-70B-AWQ & MAS & MIMIC-IV & -6.83 & [-11.57, -1.97] & 0.0496 & 0 \\
Llama3.3-70B-AWQ & AC & MedQA & +5.92 & [-1.56, 13.40] & 0.4533 & 0 \\
Llama3.3-70B-AWQ & AC & MIMIC-IV & +0.00 & [-4.05, 4.28] & 1.0000 & 0 \\
Qwen2.5-72B-AWQ & MAS & MedQA & +14.95 & [6.23, 23.68] & 0.0236 & 0 \\
Qwen2.5-72B-AWQ & MAS & MIMIC-IV & +9.26 & [4.75, 13.89] & 0.0014 & 0 \\
Qwen2.5-72B-AWQ & AC & MedQA & +13.71 & [4.98, 22.12] & 0.0315 & 0 \\
Qwen2.5-72B-AWQ & AC & MIMIC-IV & +10.19 & [5.56, 14.70] & 0.0014 & 0 \\
\bottomrule
\end{tabular}
\caption{Direct minus Native for Qwen3 and the larger backbones. CIs use case-cluster bootstrap resampling. Exploratory sign-flip $p$ values are Holm-adjusted across the 28 comparisons.}
\label{tab:new-paired}
\end{table*}

\end{document}